\documentclass[journal]{IEEEtran}
\usepackage{amsmath, amssymb}
\usepackage{graphicx}
\usepackage{cite}
\usepackage{url}
\usepackage{multirow}
\usepackage{booktabs}  
\usepackage{placeins}
\usepackage{amsmath}
\usepackage{amssymb}
\usepackage{caption}
\usepackage[hidelinks]{hyperref}

\title{FirstDiff: One-Step Diffusion-Based Anomaly Detection for Multivariate Time Series via Initial Noise Prediction}
\author{Ali Boudaghi,
Alireza Nemati,
Hadi Zare%
\thanks{Corresponding author: Ali Boudaghi (email: ali.boudaghi@ut.ac.ir).}
\thanks{Code and implementation are publicly available at: \url{https://github.com/aliramsy/FirstDiff}.}}

\begin{document}
\maketitle

\begin{abstract}
Diffusion models have recently shown strong potential for multivariate time-series anomaly detection by learning the distribution of normal data through iterative denoising. Existing diffusion-based approaches, however, typically perform anomaly detection after completing the reverse diffusion process, relying primarily on the final reconstructed signal and overlooking informative representations produced during denoising. This design incurs substantial computational cost and limits the use of intermediate diffusion information for anomaly detection.

In this paper, we propose FirstDiff, a diffusion-based anomaly detection framework based on the observation that the predicted diffusion noise at the initial reverse-diffusion evaluation already contains sufficient information for accurate anomaly detection. FirstDiff models the statistical distribution of predicted diffusion noise under normal behavior using validation data, enabling anomaly inference from a single denoising-network evaluation rather than completing the reverse diffusion trajectory.

To model complex temporal and inter-sensor dependencies, FirstDiff employs a Diffusion Transformer as the denoising backbone. Extensive experiments on five public benchmark datasets demonstrate that FirstDiff achieves state-of-the-art performance while reducing diffusion inference from the full reverse trajectory to a single denoising-network evaluation.
\end{abstract}

\section{Introduction}

Multivariate time-series anomaly detection plays a fundamental role in ensuring the reliability and safety of modern cyber-physical systems, including industrial control systems, cloud infrastructures, power grids, healthcare monitoring, and financial platforms \cite{yan2024survey}. These systems continuously generate large volumes of high-dimensional sensor measurements whose complex temporal dynamics make manual monitoring infeasible. Detecting abnormal behaviors at an early stage is therefore crucial for preventing equipment failures, identifying cyberattacks, reducing maintenance costs, and improving operational reliability \cite{park2026paano,chen2023imdiffusion}.

Existing deep learning approaches for multivariate time-series anomaly detection are generally categorized into prediction-based and reconstruction-based methods. Prediction-based approaches estimate future observations from historical measurements and identify anomalies according to prediction errors \cite{Deng_Hooi_2021,10.1016/j.knosys.2021.107757}. Reconstruction-based methods instead learn the distribution of normal data and detect anomalies through reconstruction discrepancies \cite{xu2022anomaly,Zhang2022GRELENMT}. Although these methods have achieved remarkable success, the resulting anomaly scores are still primarily derived from observable prediction or reconstruction errors. Consequently, their effectiveness may degrade when anomalies produce only subtle reconstruction discrepancies or remain close to the learned normal data manifold \cite{10.1145/3444690,10.14778/3538598.3538602}.

Recently, diffusion probabilistic models have emerged as powerful generative models capable of learning highly complex data distributions through iterative denoising \cite{ddpm}. Their remarkable success in image synthesis has inspired growing interest in diffusion-based anomaly detection \cite{bhosale2024diffusion,chen2023imdiffusion}. During denoising, the model produces a sequence of noise estimates that progressively guides a corrupted sample toward the learned data distribution. In existing diffusion-based anomaly detection methods, these intermediate estimates primarily serve the denoising procedure, while the final anomaly decision is obtained only after the complete denoising trajectory. This leaves open the question of whether the intermediate denoising information itself can support reliable anomaly detection.

Based on this observation, we propose FirstDiff, a diffusion-based anomaly detection framework that performs anomaly inference from the denoising information available at the initial reverse-diffusion evaluation. Specifically, FirstDiff characterizes normal denoising behavior by modeling the statistical distribution of the predicted diffusion noise using normal validation data. Test samples are then evaluated according to their deviation from this reference distribution. Our experiments show that the information available at this early stage is already sufficiently discriminative for accurate anomaly detection, eliminating the need to execute the remaining denoising iterations. Consequently, FirstDiff reduces diffusion inference from the complete denoising trajectory to a single network evaluation, substantially lowering the computational cost of diffusion-based anomaly detection.

To effectively model multivariate time-series, we employ a Diffusion Transformer (DiT) backbone \cite{DiT}. Unlike conventional CNN-based architectures, which primarily capture local temporal dependencies, and graph-based networks, which mainly model relationships among sensors using predefined graph structures, the Transformer employs global self-attention to jointly model dependencies across all timestamps and sensor variables within each input window. This enables the model to simultaneously capture long-range temporal patterns and complex inter-sensor interactions, making it particularly suitable for industrial multivariate time-series where anomalies often arise from intricate correlations spanning multiple sensors and distant time steps.

While FirstDiff is designed as a one-step anomaly detector, the predicted diffusion noise captures information that is complementary to conventional reconstruction errors. We therefore investigate a hybrid inference strategy that jointly exploits both representations, leading to further improvements in detection accuracy.

The main contributions of this work are summarized as follows:

\begin{itemize}
    \item \textbf{One-step diffusion-based anomaly detection.}
We introduce FirstDiff, a diffusion-based framework for multivariate time-series anomaly detection that extracts anomaly information directly from the predicted noise at the initial reverse-diffusion evaluation. Unlike conventional diffusion-based anomaly detection methods that rely on completing the reverse process to obtain a reconstructed signal, FirstDiff performs anomaly inference from a single denoising-network evaluation, substantially reducing the computational cost and inference latency of diffusion-based detection.

    \item \textbf{Predicted-noise representation for anomaly characterization.}
    We establish the predicted diffusion noise as an effective representation of normal and anomalous temporal behavior. A statistical reference distribution is estimated from normal validation data, and anomalies are identified according to the deviation of their predicted-noise representations from this reference distribution. In particular, we investigate Mahalanobis distance and other statistical measures to capture deviations while accounting for dependencies among multiple sensors.

    \item \textbf{Efficient diffusion-based inference.}
    We employ a DiT to model the temporal and inter-sensor dependencies of multivariate time-series windows. By extracting anomaly information directly from the predicted diffusion noise at the initial reverse-diffusion evaluation, the proposed framework retains the expressive modeling capability of diffusion models while avoiding the computational overhead of subsequent reverse-diffusion evaluations. Empirical inference-time measurements further demonstrate the resulting reduction in computational latency.

    \item \textbf{Comprehensive empirical evaluation and analysis.}
We evaluate FirstDiff on five widely used multivariate time-series anomaly detection benchmarks, including SMAP, MSL, SMD, SWaT, and PSM, against representative classical, deep learning, and diffusion-based methods. We further conduct ablation studies to examine different anomaly representations and statistical scoring strategies, investigate the complementarity between the proposed one-step predicted-noise signal and reconstruction-based anomaly scores, and analyze the inference latency of FirstDiff against existing diffusion-based approaches.
\end{itemize}

\section{Related Work}

\subsection{Classical Methods}

Early studies on multivariate time-series anomaly detection primarily relied on statistical analysis, signal processing techniques, and classical machine learning algorithms. Representative methods include statistical and density-based approaches such as ECOD \cite{li2022ecod}, dimensionality reduction techniques including PCA \cite{scholkopf2001estimating}, kernel-based methods such as One-Class SVM \cite{patcha2007overview}, clustering algorithms including CBLOF \cite{he2003discovering} and k-Means, ensemble-based methods such as Isolation Forest \cite{liu2008isolation}, signal decomposition approaches based on wavelet transforms \cite{kanarachos2015anomaly}, and forecasting models including ARIMA \cite{yaacob2010arima}. Owing to their simplicity, interpretability, and relatively low computational cost, these methods remain important baselines for evaluating anomaly detection algorithms. However, their underlying assumptions often limit their ability to capture the complex nonlinear relationships, high-dimensional correlations, and long-range temporal dependencies commonly observed in modern industrial multivariate time-series \cite{darban2024survey}.

\begin{figure*}[t]
    \centering
    \includegraphics[width=\textwidth]{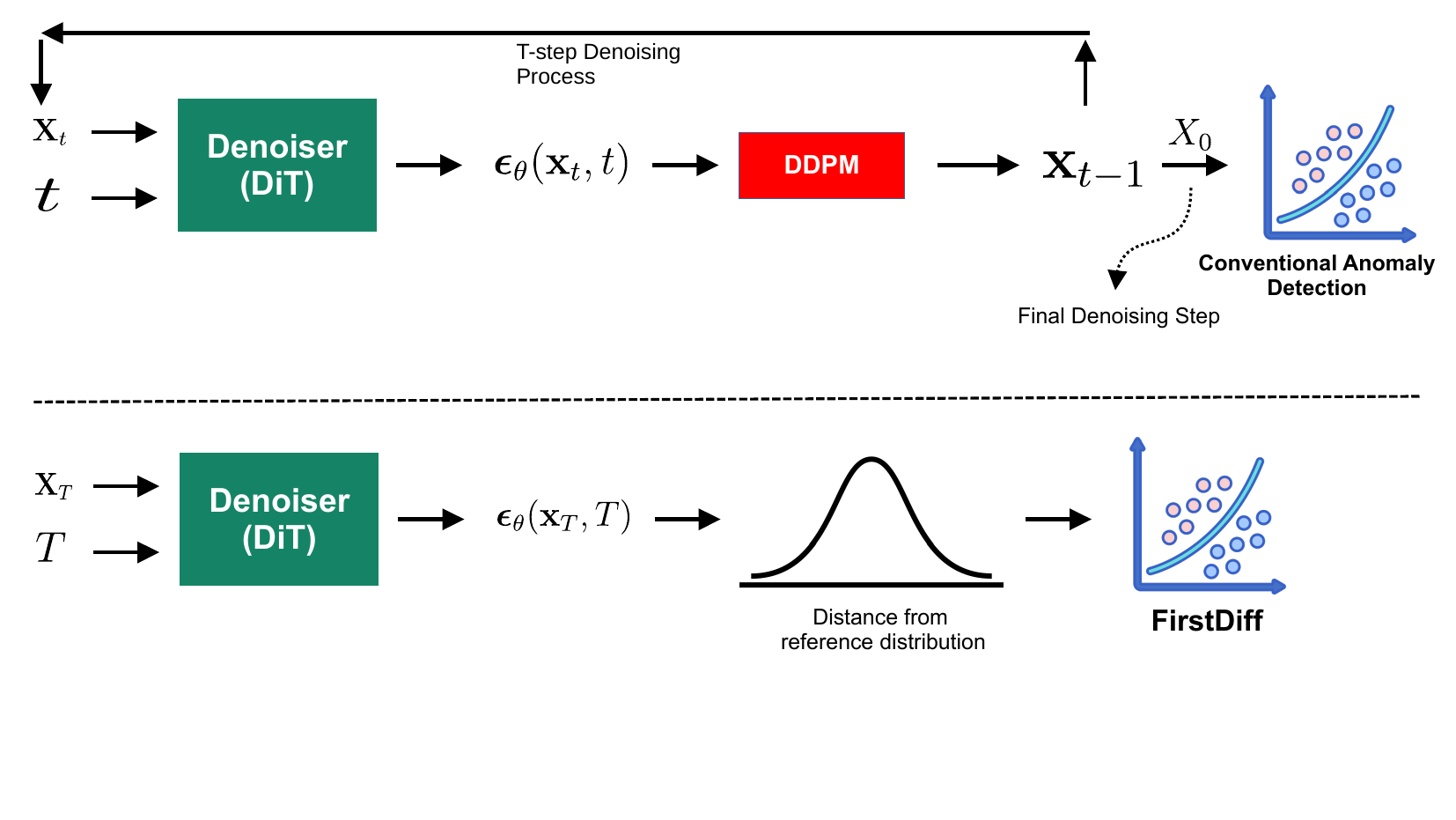}
\caption{
Overview of the proposed FirstDiff framework and its contrast with conventional diffusion-based anomaly detection. In conventional reconstruction-based inference, the noisy sample $\mathbf{X}_T$ is iteratively denoised through $T$ reverse diffusion steps to obtain the final reconstruction $\hat{\mathbf{X}}_0$, from which an anomaly score is derived. In contrast, FirstDiff uses the predicted diffusion noise $\boldsymbol{\epsilon}_{\theta}(\mathbf{X}_T,T)$ generated at the first reverse diffusion step and measures its deviation from a reference distribution estimated using normal validation data. This enables anomaly inference after a single denoising step without completing the remaining reverse diffusion trajectory.
}
    \label{fig:framework}
\end{figure*}

\begin{figure*}[t]
    \centering
    \includegraphics[width=\textwidth]{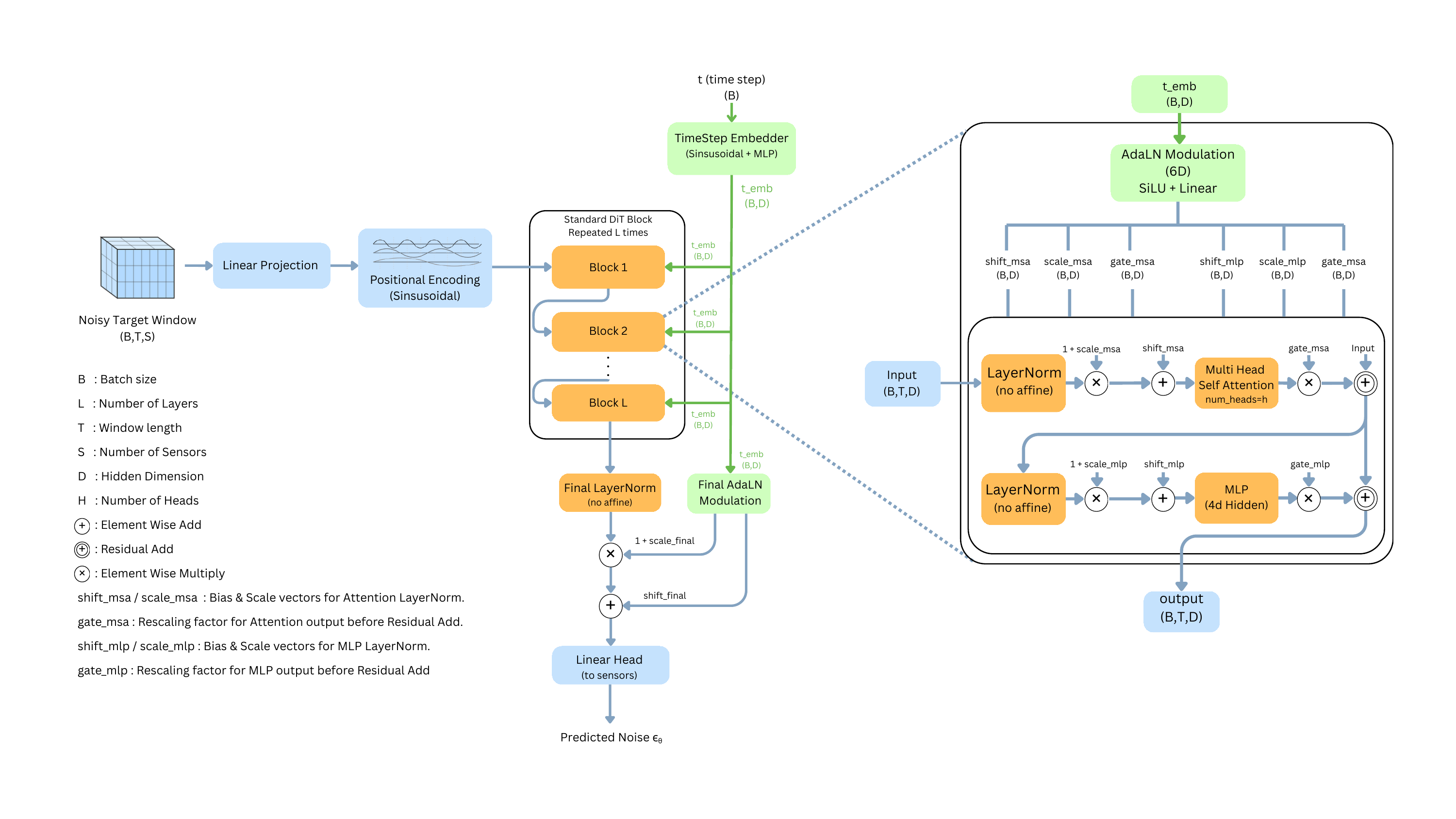}
    \caption{
Architecture of the Diffusion Transforme denoising network used in FirstDiff. A noisy multivariate time-series window is first projected into a latent embedding space and combined with positional embeddings. The embedded sequence is then processed by a stack of timestep-conditioned DiT blocks, each consisting of multi-head self-attention and an MLP with Adaptive Layer Normalization (AdaLN). Finally, a linear prediction head estimates the injected Gaussian noise, which serves as the latent representation exploited by FirstDiff for one-step anomaly detection and is also used during the reverse diffusion process for conventional reconstruction.
    }
    \label{fig:dit_architecture}
\end{figure*}

\begin{figure*}[t]
    \centering
    \includegraphics[width=\textwidth]{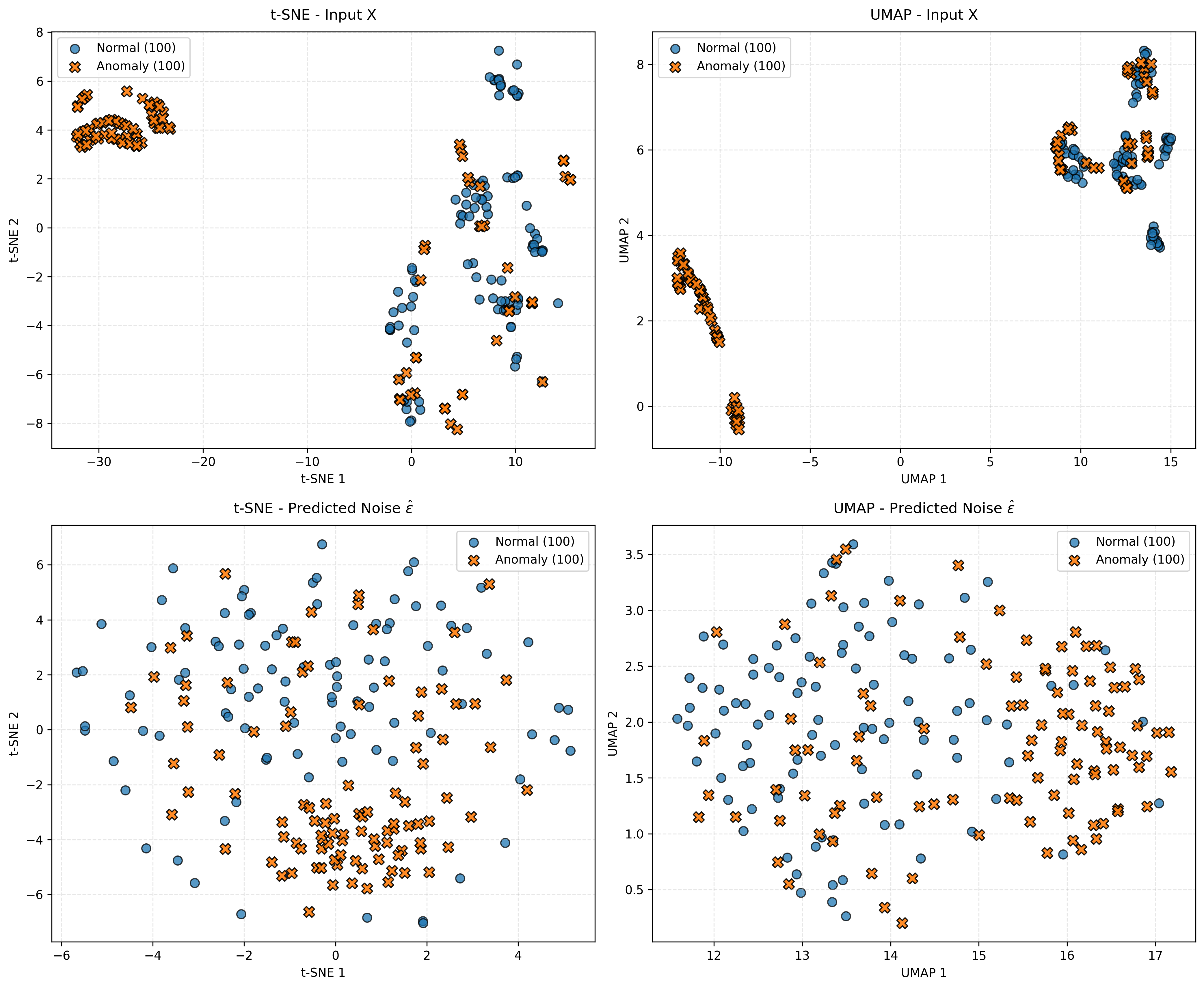}
   \caption{
Visualization of the reconstructed samples and predicted diffusion noise using t-SNE and UMAP. The first row shows the reconstructed samples $\hat{\mathbf{X}}$ obtained after the complete reverse diffusion process, while the second row shows the predicted noise $\hat{\boldsymbol{\epsilon}}$ obtained at the first reverse diffusion step. The visible separation between normal and anomalous samples in the projected spaces is provided for illustrative purposes only, as the underlying anomaly-related discrepancy arises from complex dependencies across temporal and sensor dimensions that cannot be fully represented in a two-dimensional visualization. Nevertheless, both representations exhibit distributional differences between normal and anomalous samples, indicating that the first-step predicted noise contains discriminative information that can be exploited for anomaly detection.
}
    \label{fig:latent_visualization}
\end{figure*}

\subsection{Deep Learning Methods}

The limitations of classical approaches have motivated the development of deep learning models capable of learning complex temporal dynamics directly from data. Existing deep learning methods are generally categorized into forecasting-based and reconstruction-based approaches \cite{darban2024survey}.

Forecasting-based methods identify anomalies through prediction errors between future observations and model forecasts. Representative approaches include LSTM-AD \cite{malhotra2015long}, which models temporal dynamics using recurrent neural networks, and MTAD-GAT \cite{zhao2020multivariate}, which jointly captures temporal dependencies and variable interactions through graph attention mechanisms. Although forecasting-based methods often achieve competitive detection performance, their effectiveness may deteriorate when prediction uncertainty accumulates in highly dynamic or non-stationary environments \cite{darban2024survey}.

Reconstruction-based methods instead learn the distribution of normal data and identify anomalies according to reconstruction discrepancies \cite{kingma2014autoencoding}. Representative methods include LSTM-VAE \cite{park2018multimodal}, OmniAnomaly \cite{su2019robust}, and InterFusion \cite{li2021multivariate}, which employ latent-variable models to characterize normal system behavior. More recently, transformer architectures have demonstrated remarkable success in modeling long-range temporal dependencies. TranAD \cite{tuli2022tranad} introduces self-conditioning to progressively refine anomaly detection, whereas Anomaly Transformer \cite{xu2022anomaly} exploits association discrepancy to distinguish normal and anomalous observations. In parallel, graph-based approaches explicitly model correlations among sensors using graph neural networks \cite{zhao2020multivariate}, while adversarial learning methods such as MAD-GAN \cite{li2019mad} and MSCRED \cite{mscred} improve representation learning through generative adversarial training. Despite these advances, most existing deep learning approaches ultimately derive anomaly scores from prediction or reconstruction errors.

\subsection{Diffusion-Based Anomaly Detection}

Diffusion probabilistic models have recently emerged as powerful generative models for multivariate time-series anomaly detection because of their stable optimization process and strong reconstruction capability \cite{ddpm,lin2024diffusion}. Existing diffusion-based methods generally learn the distribution of normal data and identify anomalies by measuring reconstruction errors after completing the reverse diffusion process.

Recent studies have focused primarily on improving reconstruction quality from different perspectives. ImDiffusion \cite{chen2023imdiffusion} combines masking and imputation strategies with Transformer-based architectures to better capture temporal dependencies and variable interactions. DiffAD \cite{xiao2023diffad} formulates anomaly detection as a diffusion-based imputation problem and employs guided denoising to recover corrupted observations. D$^3$R \cite{wang2023d3r} integrates time-series decomposition into the diffusion framework to mitigate the effects of distribution drift, while DiffusionAE \cite{10.1007/s10489-024-05341-0} incorporates an autoencoder to refine diffusion-based reconstruction.

Another line of research has investigated more expressive architectures for modeling dependencies within diffusion frameworks. Graph-Attention Diffusion \cite{lanko2024graph} explicitly models relationships among sensors using graph attention mechanisms, whereas ICDiffAD \cite{icdiffad2026} introduces implicit conditioning together with an SNR-guided diffusion schedule to improve reconstruction consistency during reverse diffusion. Despite these advances, the potential of the intermediate denoising process itself as a source of anomaly information remains largely unexplored.

\section{Overview and Methodology}
\label{sec:method}

\subsection{Problem Formulation and Framework Overview}

Let
\begin{equation}
\mathbf{X}
=
[\mathbf{x}_1,\mathbf{x}_2,\ldots,\mathbf{x}_L]
\in
\mathbb{R}^{L\times K},
\end{equation}
denote a multivariate time series with $L$ timestamps and $K$ sensor variables, where
$\mathbf{x}_l\in\mathbb{R}^{K}$ represents the observation at timestamp $l$. Following the unsupervised anomaly detection setting, only normal data are available during training and validation, while anomalies appear only during testing.

The input sequence is partitioned into overlapping windows of length $W$. The $i$-th window is defined as
\begin{equation}
\mathbf{X}^{(i)}
=
[\mathbf{x}_i,\mathbf{x}_{i+1},\ldots,\mathbf{x}_{i+W-1}]
\in
\mathbb{R}^{W\times K},
\end{equation}
where $W=96$ in all experiments. The objective is to learn the characteristics of normal windows and estimate the timestamp-level anomaly sequence
\begin{equation}
\hat{Y}
=
\{\hat{y}_1,\hat{y}_2,\ldots,\hat{y}_L\},
\end{equation}
where $\hat{y}_l\in\{0,1\}$ indicates whether timestamp $l$ is normal or anomalous.

Figure~\ref{fig:framework} provides an overview of the proposed FirstDiff framework in comparison with conventional reconstruction-based diffusion anomaly detection. Conventional methods complete the reverse diffusion trajectory, iteratively denoising $\mathbf{X}_T$ through $T$ steps to obtain the final reconstruction $\hat{\mathbf{X}}_0$, which is then used to derive an anomaly score. In contrast, FirstDiff performs anomaly inference directly from the predicted diffusion noise $\boldsymbol{\epsilon}_{\theta}(\mathbf{X}_T,T)$ obtained at the first reverse diffusion step. The predicted noise is compared with a reference distribution estimated from normal validation data using statistical distance measures, allowing anomaly detection without executing the remaining reverse diffusion steps. Reconstruction-based and hybrid version of our framework are investigated separately in Section~\ref{sec:Ablation}.

The denoising model used throughout the framework is illustrated in Figure~\ref{fig:dit_architecture}. We employ a DiT as the denoising backbone because its global self-attention mechanism jointly models dependencies among all sensors across all timestamps within each input window. Given a noisy multivariate window and the corresponding diffusion timestep, the DiT predicts the injected Gaussian noise required for the denoising process.

\subsection{Diffusion-Based Denoising}

The proposed framework employs the Denoising Diffusion Probabilistic Model (DDPM)~\cite{ddpm} to learn the distribution of normal multivariate time-series windows. During training, Gaussian noise is progressively added to a clean input window, while the denoising network is trained to predict the injected noise. During inference, the learned denoising process progressively removes the noise through a sequence of reverse diffusion steps, ultimately recovering a sample corresponding to the learned data distribution.

Given a clean multivariate time-series window
$\mathbf{X}_0\in\mathbb{R}^{W\times K}$,
the forward diffusion process gradually corrupts the data according to

\begin{equation}
q(\mathbf{X}_t|\mathbf{X}_{t-1})
=
\mathcal{N}
\left(
\sqrt{\alpha_t}\mathbf{X}_{t-1},
(1-\alpha_t)\mathbf{I}
\right),
\end{equation}

where $\{\alpha_t\}_{t=1}^{T}$ denotes the predefined noise schedule and $T$ is the total number of diffusion steps. Using the closed-form formulation of the forward process, the noisy sample at any diffusion step can be obtained directly as

\begin{equation}
\mathbf{X}_t
=
\sqrt{\bar{\alpha}_t}\mathbf{X}_0
+
\sqrt{1-\bar{\alpha}_t}\boldsymbol{\epsilon},
\label{eq:forward}
\end{equation}

where

\begin{equation}
\bar{\alpha}_t
=
\prod_{i=1}^{t}\alpha_i,
\end{equation}

and
$\boldsymbol{\epsilon}\sim\mathcal{N}(\mathbf{0},\mathbf{I})$
is the sampled Gaussian noise.

The DiT denoising network introduced in the previous subsection is trained to predict the injected noise by minimizing the standard DDPM objective,

\begin{equation}
\mathcal{L}
=
\mathbb{E}_{\mathbf{X}_0,t,\boldsymbol{\epsilon}}
\left[
\left\|
\boldsymbol{\epsilon}
-
\boldsymbol{\epsilon}_{\theta}(\mathbf{X}_t,t)
\right\|_2^2
\right],
\label{eq:loss}
\end{equation}

where
$\boldsymbol{\epsilon}_{\theta}(\mathbf{X}_t,t)$
denotes the predicted Gaussian noise.

During inference, the reverse diffusion process progressively removes the predicted noise to transform the noisy sample toward the learned data distribution. Each reverse diffusion step is computed as

\begin{equation}
\mathbf{X}_{t-1}
=
\frac{1}{\sqrt{\alpha_t}}
\left(
\mathbf{X}_t
-
\frac{1-\alpha_t}
{\sqrt{1-\bar{\alpha}_t}}
\boldsymbol{\epsilon}_{\theta}(\mathbf{X}_t,t)
\right)
+
\sigma_t\mathbf{z},
\label{eq:reverse}
\end{equation}

where
$\mathbf{z}\sim\mathcal{N}(\mathbf{0},\mathbf{I})$
for $t>1$ and $\mathbf{z}=\mathbf{0}$ for the final reverse step.

After completing the reverse diffusion process, diffusion-based reconstruction methods typically use the discrepancy between the input window $\mathbf{X}_0$ and the resulting sample $\hat{\mathbf{X}}_0$ as the anomaly score. A common timestamp-level reconstruction score is defined as

\begin{equation}
s_{\mathrm{rec}}(l)
=
\frac{1}{K}
\sum_{k=1}^{K}
\left|
X_0(l,k)-\hat{X}_0(l,k)
\right|.
\label{eq:recscore}
\end{equation}

Larger values indicate greater deviation between the observation and its denoised counterpart and therefore a higher likelihood of anomalous behavior.
\subsection{Predicted Noise Distribution Modeling}
\label{subsec:latent_distribution}

FirstDiff uses the predicted diffusion noise generated during the first reverse diffusion step as an anomaly indicator, rather than waiting for the complete reverse diffusion process to obtain the final output.

For a noisy sample $\mathbf{X}_t$, the denoising network predicts

\begin{equation}
\hat{\boldsymbol{\epsilon}}_t
=
\boldsymbol{\epsilon}_{\theta}(\mathbf{X}_t,t),
\end{equation}

which, under the DDPM training objective, approximates the optimal predictor

\begin{equation}
\boldsymbol{\epsilon}_{\theta}^{*}(\mathbf{X}_t,t)
=
\mathbb{E}
\left[
\boldsymbol{\epsilon}
\mid
\mathbf{X}_t
\right].
\end{equation}

As shown in Appendix~\ref{appendix:score_proof}, the optimal noise prediction is directly related to the score function of the noisy data distribution,

\begin{equation}
\nabla_{\mathbf{X}_t}
\log p_t(\mathbf{X}_t)
\approx
-
\frac{1}{\sqrt{1-\bar{\alpha}_t}}
\boldsymbol{\epsilon}_{\theta}
(\mathbf{X}_t,t),
\end{equation}

where the approximation becomes exact for the optimal predictor
$\boldsymbol{\epsilon}_{\theta}^{*}$.
Therefore, the predicted noise represents the direction that guides the sample toward regions of higher probability under the learned normal distribution. Samples drawn from normal data consequently exhibit consistent denoising directions, whereas anomalous samples generally require different correction directions and thus produce predicted noise distributions that deviate from normal denoising behavior. Figure~\ref{fig:latent_visualization} provides an illustrative visualization of the reconstructed samples and first-step predicted noise using t-SNE and UMAP. Both representations exhibit distributional differences between normal and anomalous samples, indicating that the predicted noise contains discriminative information even after a single denoising step. The anomaly detection performance of the predicted noise is further evaluated quantitatively in the main experimental results and ablation study presented in Sections~\ref{sec:Results} and~\ref{sec:Ablation}.

To model the normal predicted-noise behavior, only normal validation windows are used to estimate the predicted-noise distribution. For each validation window, Gaussian noise is added according to the forward diffusion process, and the trained DiT predicts the noise at the first reverse diffusion step,

\begin{equation}
\hat{\boldsymbol{\epsilon}}^{(1)}
=
\boldsymbol{\epsilon}_{\theta}
(\mathbf{X}_{T},T).
\end{equation}

In addition to the predicted noise, we also compute the denoising residual,

\begin{equation}
\mathbf{r}^{(1)}
=
\boldsymbol{\epsilon}
-
\hat{\boldsymbol{\epsilon}}^{(1)},
\end{equation}

which measures the prediction error of the diffusion model. For each validation window of size $W\times K$, both latent representations are reshaped into timestamp-level vectors in $\mathbb{R}^{K}$ and aggregated over the validation set. Rather than assuming that the predicted-noise or residual representations exactly follow a Gaussian distribution, we use Gaussian reference models as tractable approximations to their empirical distributions under normal operation. The corresponding statistics are estimated using the sample mean and covariance,

\[
(\boldsymbol{\mu}_{\epsilon},
\boldsymbol{\Sigma}_{\epsilon})
\qquad\text{and}\qquad
(\boldsymbol{\mu}_{r},
\boldsymbol{\Sigma}_{r}),
\]

which define the reference models of normal denoising behavior.

During inference, FirstDiff performs anomaly detection directly from the initial reverse-diffusion evaluation without executing the remaining denoising iterations. Given a latent representation
$\mathbf{z}\in\mathbb{R}^{K}$,
where $\mathbf{z}$ denotes either the predicted noise
$\hat{\boldsymbol{\epsilon}}^{(1)}$
or the residual
$\mathbf{r}^{(1)}$,
its anomaly score is computed as

\begin{equation}
s
=
d(\mathbf{z},\boldsymbol{\mu},\boldsymbol{\Sigma}),
\end{equation}

where $d(\cdot)$ measures the deviation from the learned normal predicted-noise distribution. Three complementary distance metrics are investigated:

\begin{equation}
d_{\mathrm{Euc}}
=
\|
\mathbf{z}
-
\boldsymbol{\mu}
\|_2,
\end{equation}

\begin{equation}
d_{\mathrm{Cos}}
=
1-
\frac{
\mathbf{z}^{T}\boldsymbol{\mu}
}
{
\|\mathbf{z}\|_2
\|\boldsymbol{\mu}\|_2
},
\end{equation}

and

\begin{equation}
d_{\mathrm{Mah}}
=
\sqrt{
(
\mathbf{z}
-
\boldsymbol{\mu}
)^T
\boldsymbol{\Sigma}^{-1}
(
\mathbf{z}
-
\boldsymbol{\mu}
)
},
\end{equation}

corresponding to Euclidean, cosine, and Mahalanobis distances, respectively. These metrics capture complementary notions of deviation, including magnitude, direction, and covariance-aware statistical distance.

The resulting timestamp-level predicted-noise anomaly scores are thresholded to obtain binary anomaly predictions,

\begin{equation}
\hat{y}_l=
\begin{cases}
1,& s_l>\tau,\\
0,& \text{otherwise},
\end{cases}
\end{equation}

where $\tau$ denotes the selected decision threshold.

\subsection{Hybrid Inference Strategy}

Although FirstDiff performs anomaly detection directly from the predicted diffusion noise obtained after the first reverse diffusion step, we additionally investigate whether complementary anomaly signals can be combined to improve detection performance. These signals may originate from different representations or from different statistical measures applied to the same predicted noise.

In particular, we consider combinations of the one-step predicted-noise anomaly scores with the conventional reconstruction-based anomaly score obtained after completing the reverse diffusion process. We also investigate combinations of predicted-noise scores computed using different distance measures, such as Euclidean, cosine, and Mahalanobis distances. For binary predictions, the individual anomaly decisions are combined using a simple OR rule,

\begin{equation}
\hat{y}^{\mathrm{hyb}}_l
=
\bigvee_{j=1}^{M}
\hat{y}^{(j)}_l,
\end{equation}

where $\hat{y}^{(j)}_l$ denotes the binary anomaly prediction from the $j$-th anomaly signal. Thus, a timestamp is classified as anomalous if it is identified by at least one of the combined detectors. This parameter-free strategy allows us to examine whether different anomaly signals provide complementary information.

The complete evaluation of the different hybrid combinations and their effects on detection performance is presented in the ablation study.

\section{Experiment}

\subsection{Datasets}

We evaluate the proposed framework on five widely used real-world benchmarks for multivariate time-series anomaly detection: SWaT \cite{mathur2016swat}, PSM \cite{abdulaal2021anomaly}, SMD \cite{su2019robust}, MSL \cite{hundman2018detecting}, and SMAP \cite{hundman2018detecting}. These datasets span diverse real-world applications, including industrial control systems, cloud-service monitoring, and spacecraft telemetry, and exhibit substantial variations in data dimensionality, sequence length, anomaly frequency, and system dynamics. Consequently, they provide a comprehensive benchmark for evaluating the robustness and generalization capability of anomaly detection algorithms under different operating conditions.

Specifically, SWaT is collected from a secure water treatment testbed and contains measurements from multiple sensors and actuators operating under both normal conditions and cyberattack scenarios. PSM consists of performance metrics collected from application servers at eBay and represents a large-scale cloud-service monitoring task. SMD is a server machine dataset containing operational measurements from distributed Internet services with sparse anomaly events. MSL and SMAP are two public NASA spacecraft telemetry datasets collected from the Mars Science Laboratory rover and the Soil Moisture Active Passive satellite mission, respectively. These two datasets are widely adopted as benchmark datasets for evaluating anomaly detection methods on complex multivariate telemetry signals.

\begin{table}[t]
\centering
\small
\setlength{\tabcolsep}{2pt}
\caption{Statistics of the benchmark datasets used in the experiments.}
\label{tab:datasets}
\begin{tabular}{lcccc}
\toprule
Dataset & Variables & Training Samples & Test Samples & Anomaly Ratio \\
\midrule
SWaT & 51 & 495,000 & 449,919 & 12.14\% \\
PSM  & 25 & 132,481 & 87,481  & 27.76\% \\
SMAP & 25 & 135,183 & 427,617 & 12.79\% \\
SMD  & 38 & 708,405 & 708,420 & 4.16\% \\
MSL  & 55 & 58,317  & 73,729  & 10.53\% \\
\bottomrule
\end{tabular}
\end{table}

Table~\ref{tab:datasets} summarizes the statistical characteristics of the five benchmarks. The datasets present markedly different levels of difficulty. For instance, SMD contains long multivariate sequences with a relatively low anomaly ratio (4.16\%), making anomaly localization particularly challenging, whereas PSM exhibits a considerably higher anomaly ratio (27.76\%). Moreover, the number of monitored variables ranges from 25 to 55, and the datasets originate from substantially different application domains. Evaluating on this diverse collection of benchmarks enables a comprehensive assessment of the proposed framework across a broad spectrum of anomaly patterns and operating environments.

\subsection{Evaluation Metrics}

We evaluated the proposed method using Precision, Recall, F1-score, and Average Detection Delay (ADD).
\begin{itemize}

\item \textbf{Precision (P), Recall (R), and F1-score} are defined as
\begin{equation}
\mathrm{Precision}=\frac{TP}{TP+FP},
\end{equation}
\begin{equation}
\mathrm{Recall}=\frac{TP}{TP+FN},
\end{equation}
\begin{equation}
\mathrm{F1}=
\frac{2\times \mathrm{Precision}\times \mathrm{Recall}}
{\mathrm{Precision}+\mathrm{Recall}},
\end{equation}
where $TP$, $FP$, and $FN$ denote the numbers of true positives, false positives, and false negatives, respectively. Precision measures the reliability of detected anomalies, Recall evaluates the ability to identify anomalous events, and the F1-score provides a balanced assessment of both.

\item \textbf{Average Detection Delay (ADD)} \cite{doshi2022reward} measures how quickly an anomaly is detected after its onset:
\begin{equation}
\mathrm{ADD}
=
\frac{1}{S}
\sum_{i=1}^{S}
(T_i-\rho_i),
\end{equation}
where $\rho_i$ is the starting timestamp of the $i$-th anomaly interval, $T_i$ is the first correctly detected timestamp within that interval, and $S$ is the total number of anomaly intervals. Lower ADD values indicate more timely anomaly detection.

\end{itemize}

\subsection{Baseline Methods}

We compare the proposed framework with a diverse set of representative anomaly detection methods spanning classical machine learning, deep learning, and recent diffusion-based approaches. This selection covers the major methodological paradigms in multivariate time-series anomaly detection and enables a comprehensive evaluation against both conventional and state-of-the-art techniques.

\textbf{Classical machine learning methods.} We first consider four widely adopted unsupervised anomaly detection algorithms, including PCA \cite{scholkopf2001estimating}, Isolation Forest (IForest) \cite{liu2008isolation}, CBLOF \cite{he2003discovering}, and ECOD \cite{li2022ecod}. These methods have been extensively used as standard baselines because of their simplicity, computational efficiency, and strong performance on a broad range of anomaly detection tasks.

\textbf{Deep learning methods.} To evaluate the proposed framework against modern neural-network-based approaches, we include representative methods from several categories of deep learning. GDN \cite{Deng_Hooi_2021} models dependencies among sensors through graph neural networks, while MAD-GAN \cite{li2019mad} employs adversarial learning to model normal system behavior. OmniAnomaly \cite{su2019robust} utilizes variational inference to learn latent representations of multivariate time series, whereas MSCRED \cite{mscred} reconstructs temporal correlation matrices using convolutional encoder-decoder networks. We further include TranAD \cite{tuli2022tranad}, a transformer-based anomaly detection framework that leverages self-attention to model long-range temporal dependencies.

\textbf{Diffusion-based methods.} Since our work is built upon diffusion probabilistic models, we perform extensive comparisons with the most closely related diffusion-based anomaly detection approaches. Specifically, we consider DiffAD \cite{xiao2023diffad}, which formulates anomaly detection as a diffusion-based imputation problem; ImDiffusion \cite{chen2023imdiffusion}, which integrates masking and imputation within a diffusion framework; Graph-Attention Diffusion (GAD) \cite{lanko2024graph}, which incorporates graph-attention mechanisms to model correlations among sensors during the diffusion process; and ICDIFFAD \cite{icdiffad2026}, which introduces implicit conditioning and an SNR-guided diffusion strategy to improve reverse denoising. These methods represent the closest and strongest diffusion-based competitors to the proposed framework.

For all competing methods, we use the official implementations whenever publicly available. Hyperparameters are configured according to the recommendations reported in the corresponding papers to ensure a fair comparison.

We also considered including two recent diffusion-based anomaly detection methods, namely the Diffusion Graph Model (DGM) \cite{Lan2025DiffusionGM} and the Multi-Resolution Decomposable Diffusion Model (MRDDM) \cite{Zhong2025MultiResolutionDD}. However, these methods do not provide publicly available implementations. To ensure a fair and reproducible comparison, they are therefore not included in our experimental evaluation.

\subsection{Implementation Details}

For all datasets, the multivariate time series are standardized using the statistics of the training set and segmented into overlapping windows of length $W=96$. During training, windows are extracted with a stride of 64, while validation and testing use a stride of 96. Following the unsupervised anomaly detection protocol, only normal training data are used for model optimization.

The denoising network is implemented using a DiT consisting of 6 Transformer blocks with a hidden dimension of 256 and 8 attention heads. The diffusion process employs $T=100$ diffusion timesteps with the standard linear variance schedule proposed in DDPM~\cite{ddpm}. The model is trained using the standard noise-prediction objective in Eq.~(\ref{eq:loss}) for 15 epochs with the Adam optimizer and a batch size of 128.

After training, the model parameters are fixed. The normal validation set is then used to estimate the Gaussian statistics of the predicted diffusion noise described in Section~\ref{subsec:latent_distribution}. During inference, FirstDiff performs anomaly detection from the predicted noise obtained at the first reverse diffusion step. Reconstruction-based and hybrid variants are additionally evaluated to investigate the effectiveness and complementarity of the proposed one-step anomaly signal.

All experiments are conducted on a workstation equipped with an NVIDIA GeForce RTX 4090 GPU.

\begin{table*}[t]
\centering
\caption{Performance comparison of different models on five anomaly detection benchmarks.}
\label{tab}
\resizebox{\textwidth}{!}{
\begin{tabular}{l|ccc|ccc|ccc|ccc|ccc|c}
\toprule
\multirow{2}{*}{\textbf{Model}}
& \multicolumn{3}{c|}{\textbf{SMD}}
& \multicolumn{3}{c|}{\textbf{PSM}}
& \multicolumn{3}{c|}{\textbf{MSL}}
& \multicolumn{3}{c|}{\textbf{SWaT}}
& \multicolumn{3}{c|}{\textbf{SMAP}}
& \multirow{2}{*}{\textbf{Avg. F1}} \\
\cline{2-16}
& P & R & F1
& P & R & F1
& P & R & F1
& P & R & F1
& P & R & F1
& \\
\midrule

CBLOF
& 21.87 & 79.93 & 34.34
& 59.46 & 92.25 & 72.30
& 16.90 & 27.96 & 21.06
& 15.75 & \underline{93.32} & 26.95
& 46.86 & 60.65 & 52.86
& 41.50 \\

ECOD
& 18.69 & 50.16 & 27.23
& 86.89 & 78.72 & 82.61
& 20.71 & 31.59 & 25.02
& 21.30 & 93.02 & 34.66
& 44.76 & 53.58 & 48.78
& 43.66 \\

Isolation Forest
& 15.92 & 59.21 & 25.10
& 91.90 & 80.96 & 86.06
& 20.09 & 33.28 & 25.06
& 37.00 & 89.47 & 52.24
& 44.95 & 61.99 & 52.10
& 48.11 \\

PCA
& 20.09 & 78.06 & 31.96
& 84.62 & 82.59 & 83.59
& 16.60 & 27.96 & 20.83
& 16.83 & 93.04 & 28.50
& 52.33 & 48.30 & 50.24
& 43.02 \\

OmniAnomaly
& 65.24 & 44.09 & 52.62
& \textbf{99.99} & 72.71 & 84.06
& 87.85 & 67.23 & 75.53
& 50.19 & 86.36 & 63.47
& 91.63 & 55.24 & 68.93
& 68.92 \\

GDN
& 51.70 & 26.00 & 34.59
& 43.17 & 66.45 & 51.96
& 88.91 & 52.77 & 66.22
& 48.31 & 86.73 & 61.94
& 92.17 & 55.98 & 69.66
& 56.88 \\

MAD\_GAN
& 64.50 & 69.10 & 66.59
& 78.01 & 85.85 & 81.74
& 85.73 & 64.80 & 73.44
& 22.82 & 90.01 & 36.41
& 92.12 & 56.42 & 69.98
& 65.63 \\

MSCRED
& 42.58 & 31.10 & 35.83
& 50.79 & 89.01 & 64.48
& 83.82 & 65.66 & 73.58
& 23.33 & 90.01 & 37.06
& 91.98 & 58.57 & 71.50
& 56.49 \\

TranAD
& 60.60 & \textbf{82.67} & 69.92
& 53.23 & 94.10 & 67.81
& 87.60 & 32.02 & 46.90
& 21.62 & 90.01 & 34.87
& 92.54 & 64.38 & 75.93
& 59.09 \\

Imdiffusion
& \underline{87.29} & 77.11 & \underline{81.77}
& 98.03 & 94.48 & \underline{96.20}
& 87.74 & 84.12 & 85.84
& \textbf{99.92} & 66.61 & 79.93
& 87.24 & 92.19 & 89.63
& 86.67 \\

DiffAD
& \textbf{95.86} & 47.61 & 63.58
& 96.31 & \underline{95.12} & 95.71
& 89.46 & \underline{91.81} & \underline{90.61}
& 38.23 & 81.81 & 52.10
& \underline{95.95} & 94.87 & \underline{95.38}
& 79.48 \\

GAD
& 82.69 & 81.24 & \textbf{81.91}
& \underline{99.75} & 88.58 & 93.83
& \underline{90.95} & 85.81 & 88.30
& \underline{95.87} & 78.78 & \underline{86.46}
& 81.33 & \textbf{97.29} & 88.60
& \underline{87.82} \\

FirstDiff (Mah($\epsilon$)) (\textbf{Ours})
& 81.47 & \underline{81.48} & 81.48
& 97.11 & \textbf{97.42} & \textbf{97.26}
& \textbf{92.43} & \textbf{93.59} & \textbf{93.01}
& 95.40 & \textbf{97.96} & \textbf{96.66}
& \textbf{96.59} & \underline{95.98} & \textbf{96.28}
& \textbf{92.94} \\

\bottomrule
\end{tabular}
}
\end{table*}

\begin{table}[t]
\centering
\caption{Comparison of anomaly detection models in terms of Average Detection Delay on five anomaly detection benchmarks.}
\label{tab:model_add}

\renewcommand{\arraystretch}{0.85}
\small

\resizebox{\columnwidth}{!}{%
\begin{tabular}{l|ccccc}
\toprule
\textbf{Method}
& \textbf{SMD}
& \textbf{PSM}
& \textbf{MSL}
& \textbf{SWaT}
& \textbf{SMAP} \\
\midrule

CBLOF
& \textbf{24.05}
& 85.99
& 174.47
& \underline{129.59}
& 419.95 \\

ECOD
& 51.38
& 181.60
& 165.00
& 153.83
& 404.00 \\

Isolation Forest
& 42.96
& 150.62
& 162.13
& 294.50
& 407.68 \\

PCA
& \underline{26.06}
& 142.42
& 174.47
& 140.20
& 463.91 \\

OmniAnomaly
& 61.17
& 199.72
& 115.99
& 392.23
& 380.10 \\

GDN
& 71.25
& 157.40
& 116.79
& 397.24
& 379.23 \\

MAD\_GAN
& 40.84
& 161.69
& 113.23
& 299.48
& 374.35 \\

TranAD
& 29.75
& 88.35
& 160.86
& 281.31
& 322.34 \\

MSCRED
& 70.39
& 114.19
& 106.09
& 331.77
& 414.63 \\

ImDiffusion
& 36.89
& 52.21
& \underline{62.59}
& 646.54
& \textbf{133.38} \\

DiffAD
& 64.27
& \underline{40.75}
& \textbf{41.44}
& 1430.66
& 214.74 \\

GAD
& 29.29
& 116.05
& 75.06
& 450.74
& 143.65 \\

FirstDiff
& 35.90
& \textbf{26.17}
& 65.36
& \textbf{110.14}
& \underline{142.06} \\

\bottomrule
\end{tabular}%
}
\end{table}

\section{Results And Discussion}
\label{sec:Results}

\subsection{Overall Anomaly Detection Performance}

Table~\ref{tab} reports the precision, recall, and F1-score of FirstDiff and the compared anomaly detection methods on five widely used multivariate time-series benchmarks. Overall, FirstDiff achieves the highest average F1-score of 92.94\%, outperforming all competing methods, including the previous best-performing GAD model with an average F1-score of 87.82\%. This result demonstrates that modeling the predicted diffusion noise after only the first reverse diffusion step provides a highly effective anomaly signal despite avoiding the complete diffusion trajectory.

FirstDiff achieves particularly strong performance on PSM, MSL, SWaT, and SMAP. On MSL, we obtain an F1-score of 93.01\%, which is the highest among all evaluated methods, together with a precision of 92.43\% and recall of 93.59\%. On SWaT, FirstDiff achieves the highest F1-score of 96.66\%, substantially exceeding the 86.46\% obtained by GAD. The corresponding recall of 97.96\% also indicates that FirstDiff successfully identifies a large proportion of anomalous observations while maintaining high precision.

On PSM, FirstDiff obtains an F1-score of 97.26\%, closely outperforming ImDiffusion 96.20\% and DiffAD 95.71\%. On SMAP, FirstDiff achieves an F1-score of 96.28\%, again exceeding all compared methods, with precision and recall of 96.59\% and 95.98\%, respectively. On SMD, FirstDiff obtains an F1-score of 81.48\%, which is competitive with GAD 81.91\% and ImDiffusion 81.77\%. Although FirstDiff does not achieve the highest F1-score on SMD, its overall performance across the five datasets remains substantially stronger, as reflected by its highest average F1-score.

These results are particularly notable because FirstDiff performs anomaly inference after only one reverse diffusion step. In contrast, diffusion-based approaches such as ImDiffusion, GAD and DiffAD rely on the full denoising process for their anomaly inference. The results therefore indicate that the information required for effective anomaly discrimination can be extracted from the predicted diffusion noise at the earliest stage of the reverse process, rather than requiring completion of the entire trajectory.

\subsection{Detection Delay}

Table~\ref{tab:model_add} reports the Average Detection Delay (ADD), which measures how quickly an anomaly is detected after its onset. Lower ADD values indicate earlier detection and therefore better temporal responsiveness.

FirstDiff achieves particularly strong detection-delay performance on PSM and SWaT. On PSM, FirstDiff obtains the lowest ADD among all compared methods, with a value of 26.17, outperforming DiffAD and ImDiffusion. This result indicates that the first-step predicted-noise representation can identify anomalous behavior substantially earlier than the competing diffusion-based methods. On SWaT, FirstDiff achieves an ADD of 110.14, which is also the lowest among all compared methods. In particular, it substantially reduces the detection delay compared with GAD, ImDiffusion and DiffAD.

On MSL, FirstDiff obtains an ADD of 65.36, which is close to the best result of ImDiffusion (62.59) and lower than most of the other compared methods. On SMAP, FirstDiff achieves an ADD of 142.06, which is competitive with GAD (143.65) and substantially lower than DiffAD (214.74), TranAD (322.34), and several other methods.

The results on SMD are comparatively less favorable. FirstDiff obtains an ADD of 35.90, whereas CBLOF and PCA achieve lower delays of 24.05 and 26.06, respectively. Nevertheless, the difference is relatively modest, and FirstDiff maintains strong overall detection performance on SMD as reflected by its F1-score.

Overall, FirstDiff achieves the lowest ADD on two of the five datasets, PSM and SWaT, and remains competitive on the remaining benchmarks. These results provide additional evidence that the predicted diffusion noise obtained after only the first reverse diffusion step contains sufficient information not only for accurate anomaly discrimination, but also for reducing anomaly detection delay.

\subsection{Inference Time Analysis}

\begin{table*}[t]
\centering
\caption{Inference time (ms) per test window of length 96 across five anomaly detection benchmarks.}
\label{tab:inference_time}

\renewcommand{\arraystretch}{0.9}
\small

\begin{tabular}{l|ccccc|c}
\hline
\textbf{Method} & \textbf{SMD} & \textbf{SMAP} & \textbf{SWaT} & \textbf{MSL} & \textbf{PSM} & \textbf{Average} \\
\hline

Imdiffusion &
204.04 & 202.95 & 205.44 & 202.78 & 207.52 & 204.55 \\

DiffAD &
272.90 & 269.33 & 269.02 & 266.37 & 264.07 & 268.34 \\

GAD &
370.71 & 370.89 & 374.01 & 370.56 & 374.39 & 372.91 \\

FirstDiff (Full-step) &
99.05 & 99.15 & 101.21 & 98.95 & 100.35 & 99.74 \\

FirstDiff (One-step) &
\textbf{1.16} & \textbf{1.15} & \textbf{1.18} & \textbf{1.60} & \textbf{1.16} & \textbf{1.25} \\

\hline
\end{tabular}
\end{table*}

To evaluate the computational efficiency of the proposed approach, we measure the inference time required to process a single test window of length 96. All measurements are performed on the same hardware using GPU synchronization, with 10 warm-up runs followed by 100 timed runs. For diffusion-based methods, the reported inference time includes the complete reverse-diffusion procedure. FirstDiff is evaluated in two modes: the proposed one-step Mahalanobis inference and the full-step reconstruction-based inference. The results are reported in Table~\ref{tab:inference_time}.

As shown in Table~\ref{tab:inference_time}, the one-step variant of FirstDiff provides a substantial reduction in inference time compared with diffusion-based reconstruction approaches. Across the five datasets, FirstDiff requires only 1.12--1.60~ms per window for one-step Mahalanobis inference, with an average inference time of approximately 1.23~ms. In contrast, its full-step reconstruction requires approximately 99--101~ms per window, corresponding to an approximately 80$\times$ increase in inference time.

More importantly, the one-step FirstDiff variant is substantially faster than all compared diffusion-based reconstruction methods. While Imdiffusion, DiffAD, and GAD require approximately 203--208~ms, 264--273~ms, and 371--374~ms per window, respectively, FirstDiff requires only approximately 1.23~ms in its one-step configuration. Thus, the proposed one-step formulation reduces the computational cost by approximately two orders of magnitude compared with the existing diffusion-based approaches, while avoiding the need to execute the complete reverse-diffusion trajectory during inference.

The full-step reconstruction version of FirstDiff is also computationally more efficient than the compared diffusion-based methods. Its average inference time is approximately 99.74~ms per window, compared with approximately 204.55~ms for Imdiffusion, 268.34~ms for DiffAD, and 372.91~ms for GAD. This corresponds to reductions of approximately 51.2\%, 62.8\%, and 73.2\%, respectively. These results demonstrate that FirstDiff provides computational advantages not only through its one-step inference strategy, but also when the complete reconstruction-based diffusion process is used.

\section{Ablation Study}
\label{sec:Ablation}

\begin{table*}[t]
\centering
\caption{Performance comparison of FirstDiff variants on five anomaly detection benchmarks.}\label{tab:FirstDiff_variants}
\resizebox{\textwidth}{!}{%
\begin{tabular}{l|cc|ccccc|c}
\toprule
\textbf{Variant}
& \textbf{One-step Inference}
& \textbf{Hybrid}
& \textbf{SMD}
& \textbf{PSM}
& \textbf{MSL}
& \textbf{SWaT}
& \textbf{SMAP}
& \textbf{Avg. F1} \\
\midrule

Reconstruction
& No & No
& \underline{90.53}
& 96.36
& 88.41
& 89.47
& 76.85
& 88.32 \\

$\epsilon$
& Yes & No
& 79.04
& 97.19
& 91.41
& 96.27
& \underline{96.44}
& 92.07 \\

Residual
& Yes & No
& 88.97
& 98.14
& 88.25
& 90.37
& 93.16
& 91.78 \\

Mah($\epsilon$)
& Yes & No
& 81.48
& 97.26
& 93.01
& 96.66
& 96.28
& 92.94 \\

Mah(Residual)
& Yes & No
& 83.52
& 95.72
& 88.77
& 92.08
& 93.11
& 90.64 \\

Cos($\epsilon$)
& Yes & No
& 82.17
& 98.30
& 89.26
& 93.79
& 96.25
& 91.95 \\

Cos(Residual)
& Yes & No
& 48.08
& 90.61
& 91.31
& 66.86
& 96.18
& 78.61 \\

Reconstruction+$\epsilon$
& No & Yes
& \underline{90.53}
& \textbf{98.53}
& 92.41
& 96.45
& 87.97
& 93.18 \\

Reconstruction+Residual
& No & Yes
& \textbf{90.68}
& 98.02
& 88.57
& 90.38
& 87.37
& 91.00 \\

Reconstruction+Mah(Residual)
& No & Yes
& \underline{90.53}
& 96.98
& 88.77
& 92.16
& 87.29
& 91.15 \\

Reconstruction+Mah($\epsilon$)
& No & Yes
& \underline{90.53}
& \underline{98.34}
& \textbf{94.20}
& 96.59
& 88.03
& \underline{93.54} \\

Reconstruction+Cos(Residual)
& No & Yes
& 82.98
& 94.03
& 92.28
& 77.74
& 88.29
& 87.06 \\

Reconstruction+Cos($\epsilon$)
& No & Yes
& \underline{90.53}
& \textbf{98.53}
& 90.29
& 96.37
& 88.24
& 92.79 \\

Cos($\epsilon$)+Mah($\epsilon$)
& Yes & Yes
& 83.49
& 98.30
& 93.00
& \textbf{97.26}
& 96.30
& \textbf{93.67} \\

$\epsilon$+Mah($\epsilon$)
& Yes & Yes
& 81.60
& 97.26
& \underline{93.76}
& \underline{96.83}
& \underline{96.44}
& 93.18 \\

$\epsilon$+Cos($\epsilon$)
& Yes & Yes
& 82.93
& 98.30
& 92.23
& 96.76
& \textbf{96.65}
& 93.37 \\

\bottomrule
\end{tabular}
}
\end{table*}

We conduct an ablation study to investigate the contribution of the one-step denoising signal, the choice of anomaly representation and statistical distance, and the effect of combining complementary anomaly signals. Table~\ref{tab:FirstDiff_variants} reports the F1-score of the evaluated variants across the five benchmark datasets. The columns \emph{One-step} and \emph{Hybrid} indicate whether the variant relies on one-step inference and combines multiple anomaly signals respectively.

\subsection{Effect of One-Step Inference}

We first examine whether anomaly detection can be performed effectively from the first reverse diffusion step rather than after completing the full denoising trajectory. The reconstruction-based variant, which requires all reverse diffusion steps, achieves an average F1-score of 88.32\%. In contrast, the one-step variants consistently achieve strong performance, with average F1-scores ranging from 78.61\% to 92.94\% across the different representations and statistical distance measures evaluated in Table~\ref{tab:FirstDiff_variants}. In particular, several one-step configurations substantially outperform the reconstruction-based variant, demonstrating that completing the reverse diffusion process is not necessary to obtain a discriminative anomaly signal.

The advantage of one-step inference is particularly evident on PSM, MSL, SWaT, and SMAP, where the best one-step variants achieve F1-scores of 98.30\%, 93.01\%, 97.26\%, and 96.65\%, respectively, compared with 96.36\%, 88.41\%, 89.47\%, and 76.85\% obtained by the reconstruction-based variant. On SMD, the reconstruction-based variant achieves 90.53\%, while the strongest one-step configuration reaches 83.52\%. Thus, although the benefit of one-step inference is dataset-dependent, the results demonstrate that informative anomaly signals are already available at the beginning of the denoising trajectory.

\subsection{Effect of Statistical Distance}

We next examine the effect of the statistical distance used to measure deviations of the predicted diffusion noise from its learned normal distribution. We evaluate Euclidean distance, cosine distance, and Mahalanobis distance, denoted by $\epsilon$, $\mathrm{Cos}(\epsilon)$, and $\mathrm{Mah}(\epsilon)$, respectively. Their average F1-scores are 92.07\%, 91.95\%, and 92.94\%, respectively.

Mahalanobis distance achieves the strongest average performance among the individual predicted-noise variants. It obtains F1-scores of 93.01\% on MSL and 96.66\% on SWaT, while also maintaining competitive performance on the remaining datasets. Cosine distance performs particularly well on PSM, where it reaches 98.30\%, and achieves 96.25\% and 96.18\% on SMAP for the predicted-noise and residual representations, respectively. The residual-based cosine distance, however, performs substantially worse on SMD and SWaT, illustrating that the effectiveness of a distance measure depends on the representation to which it is applied.

Overall, the results indicate that accounting for the covariance structure of the predicted-noise distribution is beneficial for the proposed statistical formulation. Mahalanobis distance therefore provides the strongest single-signal configuration, achieving an average F1-score of 92.94\%, and is adopted as the primary FirstDiff configuration.

\subsection{Effect of Hybrid Anomaly Signals}

Finally, we investigate whether combining anomaly signals can provide complementary information beyond individual detectors. The hybrid configurations include both combinations of reconstruction-based and one-step signals and combinations of different statistical measures applied to the predicted noise.

Combining reconstruction with the predicted-noise signal improves the average F1-score from 92.07\% for $\epsilon$ alone to 93.18\% for Reconstruction+$\epsilon$. The strongest reconstruction-based combination is Reconstruction+Mah($\epsilon$), which achieves an average F1-score of 93.54\%. This configuration also obtains the highest F1-score on MSL, reaching 94.20\%. These results indicate that the information captured by the first-step predicted noise is complementary to the information contained in the final reconstruction.

Complementarity is also observed when different statistical distances are applied to the same predicted-noise representation. Cos($\epsilon$)+Mah($\epsilon$) achieves the highest average F1-score among all evaluated variants, reaching 93.67\%. It also obtains the best result on SWaT, with an F1-score of 97.26\%. The $\epsilon$+Cos($\epsilon$) and $\epsilon$+Mah($\epsilon$) combinations achieve average F1-scores of 93.37\% and 93.18\%, respectively, further indicating that different distance measures capture complementary aspects of the predicted-noise distribution.

These results show that additional gains can be obtained by combining anomaly signals, either across different stages of the diffusion process or across different statistical views of the same predicted-noise representation. However, these hybrid configurations are considered extensions of the core framework rather than the primary FirstDiff formulation. The main contribution of FirstDiff remains one-step anomaly detection using the predicted diffusion noise and its learned normal distribution.

\section{Conclusion}

In this paper, we introduced FirstDiff, a diffusion-based framework for multivariate time-series anomaly detection that exploits predicted diffusion noise generated during the denoising process rather than relying solely on the final reconstructed signal. The key finding of this work is that the predicted noise obtained at the initial reverse-diffusion evaluation already provides a highly informative representation for effective anomaly detection. We further evaluated predicted-noise representations at multiple diffusion timesteps and found that their detection performance remained relatively stable, with later timestep representations providing no substantial improvement over the initial prediction. We therefore adopt the initial predicted-noise representation in FirstDiff, as it provides comparable detection performance while requiring only a single denoising-network evaluation. By modeling its statistical distribution using normal validation data and measuring deviations from this reference model, FirstDiff enables efficient anomaly inference without completing the reverse-diffusion trajectory.

We further investigated different representations and statistical distance measures for one-step anomaly detection. The experimental results show that the predicted diffusion noise provides a strong anomaly signal across diverse benchmark datasets, with Mahalanobis distance providing particularly robust performance. Importantly, the results also demonstrate that the proposed one-step signal is not merely a proxy for the final diffusion output, but provides complementary information that can be exploited alongside conventional reconstruction-based anomaly scores. Our ablation study further examines different combinations of anomaly signals and distance measures, showing that combining complementary cues can provide additional performance gains.

Beyond the proposed one-step inference mechanism, the reconstruction-based variant of FirstDiff also achieves competitive performance against several established baseline methods. Moreover, the inference-time analysis shows that FirstDiff remains computationally efficient even when the complete diffusion trajectory is used, while the proposed one-step formulation provides a substantially larger reduction in inference latency. These results indicate that the Diffusion Transformer denoising backbone can effectively capture complex temporal and inter-sensor dependencies in multivariate time series while supporting efficient diffusion-based anomaly detection.

Overall, FirstDiff provides a different perspective on diffusion-based anomaly detection by shifting the focus from the final diffusion output to the information available at the earliest stage of denoising. Future work could investigate whether similar early-step anomaly signals can be exploited with alternative diffusion formulations, adaptive selection of denoising steps, and more efficient distribution modeling for large-scale and online multivariate time-series anomaly detection.

\bibliographystyle{IEEEtran}
\bibliography{references}

\clearpage
\appendix
\section{Appendix}

\subsection{Relationship Between Noise Prediction and the Score Function}
\label{appendix:score_proof}

This appendix derives the relationship between the diffusion noise predictor and the score function used in Section~\ref{sec:method}. Specifically, we show that

\begin{equation}
\nabla_{\mathbf{X}_t}\log p_t(\mathbf{X}_t)
=
-
\frac{1}{\sqrt{1-\bar{\alpha}_t}}
\boldsymbol{\epsilon}_{\theta}(\mathbf{X}_t,t),
\end{equation}

where
$\boldsymbol{\epsilon}_{\theta}(\mathbf{X}_t,t)$
denotes the optimal diffusion noise predictor.

Starting from the forward diffusion process,

\begin{equation}
\mathbf{X}_t
=
\sqrt{\bar{\alpha}_t}\mathbf{X}_0
+
\sqrt{1-\bar{\alpha}_t}\boldsymbol{\epsilon},
\qquad
\boldsymbol{\epsilon}
\sim
\mathcal{N}(\mathbf{0},\mathbf{I}),
\label{eq:forward_appendix}
\end{equation}

the conditional distribution of the noisy sample is

\begin{equation}
p(\mathbf{X}_t|\mathbf{X}_0)
=
\mathcal{N}
\left(
\sqrt{\bar{\alpha}_t}\mathbf{X}_0,
(1-\bar{\alpha}_t)\mathbf{I}
\right).
\end{equation}

The logarithm of this Gaussian distribution is

\begin{equation}
\log
p(\mathbf{X}_t|\mathbf{X}_0)
=
-\frac{1}{2(1-\bar{\alpha}_t)}
\left\|
\mathbf{X}_t
-
\sqrt{\bar{\alpha}_t}\mathbf{X}_0
\right\|_2^2
+C,
\end{equation}

where $C$ is independent of $\mathbf{X}_t$.

Differentiating with respect to $\mathbf{X}_t$ gives

\begin{equation}
\nabla_{\mathbf{X}_t}
\log
p(\mathbf{X}_t|\mathbf{X}_0)
=
-
\frac{
\mathbf{X}_t
-
\sqrt{\bar{\alpha}_t}\mathbf{X}_0
}
{1-\bar{\alpha}_t}.
\end{equation}

Using the forward diffusion equation (\ref{eq:forward_appendix}),

\begin{equation}
\mathbf{X}_t
-
\sqrt{\bar{\alpha}_t}\mathbf{X}_0
=
\sqrt{1-\bar{\alpha}_t}
\,\boldsymbol{\epsilon},
\end{equation}

we obtain

\begin{equation}
\nabla_{\mathbf{X}_t}
\log
p(\mathbf{X}_t|\mathbf{X}_0)
=
-
\frac{
\boldsymbol{\epsilon}
}
{\sqrt{1-\bar{\alpha}_t}}.
\label{eq:conditional_score}
\end{equation}

The unconditional score function is

\begin{equation}
\nabla_{\mathbf{X}_t}
\log
p_t(\mathbf{X}_t)
=
\mathbb{E}
\left[
\nabla_{\mathbf{X}_t}
\log
p(\mathbf{X}_t|\mathbf{X}_0)
\,\middle|\,
\mathbf{X}_t
\right],
\end{equation}

which follows from Fisher's identity. Substituting
(\ref{eq:conditional_score}) yields

\begin{equation}
\nabla_{\mathbf{X}_t}
\log
p_t(\mathbf{X}_t)
=
-
\frac{
\mathbb{E}
[
\boldsymbol{\epsilon}
|
\mathbf{X}_t
]
}
{\sqrt{1-\bar{\alpha}_t}}.
\end{equation}

In DDPM, the denoising network is trained by minimizing

\begin{equation}
\mathcal{L}
=
\mathbb{E}
\left[
\|
\boldsymbol{\epsilon}
-
\boldsymbol{\epsilon}_{\theta}
(\mathbf{X}_t,t)
\|_2^2
\right].
\end{equation}

The optimal solution of this least-squares objective is the conditional expectation,

\begin{equation}
\boldsymbol{\epsilon}_{\theta}^{*}
(\mathbf{X}_t,t)
=
\mathbb{E}
[
\boldsymbol{\epsilon}
|
\mathbf{X}_t
].
\end{equation}

Replacing the conditional expectation with the optimal predictor gives

\begin{equation}
\nabla_{\mathbf{X}_t}
\log
p_t(\mathbf{X}_t)
=
-
\frac{
\boldsymbol{\epsilon}_{\theta}^{*}
(\mathbf{X}_t,t)
}
{\sqrt{1-\bar{\alpha}_t}}.
\end{equation}

In practice, the trained denoising network approximates the optimal predictor, yielding

\begin{equation}
\nabla_{\mathbf{X}_t}
\log
p_t(\mathbf{X}_t)
\approx
-
\frac{
\boldsymbol{\epsilon}_{\theta}
(\mathbf{X}_t,t)
}
{\sqrt{1-\bar{\alpha}_t}},
\end{equation}

which is the relationship used throughout this work.

\end{document}